\documentclass[conference]{IEEEtran}

\usepackage{url}
\usepackage{booktabs}
\usepackage{amsmath}
\usepackage{amsfonts}
\usepackage{bm}

\usepackage{caption}
\usepackage{subcaption}

\usepackage{graphicx}

\usepackage{orcidlink}

\usepackage{xspace}
\newcommand{\ie}{i.\,e.,\xspace}
\newcommand{\eg}{e.\,g.,\xspace}

\usepackage[backend=biber,style=ieee,natbib=true]{biblatex} 
\begin{document}

\title{HyperAggregation: Aggregating over Graph Edges with Hypernetworks} 

\author{\IEEEauthorblockN{Nicolas Lell\orcidlink{0000-0002-6079-6480}}
\IEEEauthorblockA{Ulm University\\
Ulm, Germany\\
Email: nicolas.lell@uni-ulm.de}
\and
\IEEEauthorblockN{Ansgar Scherp\orcidlink{0000-0002-2653-9245}}
\IEEEauthorblockA{Ulm University\\
Ulm, Germany\\
Email: ansgar.scherp@uni-ulm.de}}

\maketitle

\begin{abstract}
HyperAggregation is a hypernetwork-based aggregation function for Graph Neural Networks.
It uses a hypernetwork to dynamically generate weights in the size of the current neighborhood, which are then used to aggregate this neighborhood. 
This aggregation with the generated weights is done like an MLP-Mixer channel mixing over variable-sized vertex neighborhoods.
We demonstrate HyperAggregation in two models, \textit{GraphHyperMixer} is a model based on MLP-Mixer while \textit{GraphHyperConv} is derived from a GCN but with a hypernetwork-based aggregation function. 
We perform experiments on diverse benchmark datasets for the vertex classification, graph classification, and graph regression tasks.
The results show that HyperAggregation can be effectively used for homophilic and heterophilic datasets in both inductive and transductive settings.
GraphHyperConv performs better than GraphHyperMixer and is especially strong in the transductive setting.
On the heterophilic dataset Roman-Empire it reaches a new state of the art.
On the graph-level tasks our models perform in line with similarly sized models.
Ablation studies investigate the robustness against various hyperparameter choices.

The implementation of HyperAggregation as well code to reproduce all experiments is available under \url{https://github.com/Foisunt/HyperAggregation}.

\end{abstract}

\IEEEpeerreviewmaketitle

\section{Introduction}

Graph Neural Networks (GNNs) are a powerful mechanism to learn representations of graph data~\citep{DBLP:series/synthesis/2020Hamilton}.
One of the challenges of utilizing graphs is the irregular nature of the graph structure, \ie the varying number of neighboring vertices.
Common message-passing GNNs aggregate neighborhood information through a weighted average of the embeddings of neighboring vertices~\citep{DBLP:conf/iclr/KipfW17,DBLP:conf/iclr/VelickovicCCRLB18}.
A hypernetwork is a network that learns to predict the weights of another network, which is called the \textit{target network}~\citep{DBLP:conf/iclr/HaDL17}.
Recently, \citet{DBLP:conf/acl/MaiPFCMF023} proposed the HyperMixer, an MLP-Mixer-style~\citep{DBLP:conf/nips/TolstikhinHKBZU21} model for Natural Language Processing (NLP) that utilizes hypernetworks to mix information between tokens.
The HyperMixer is an efficient alternative to the self-attention mechanism in transformers.

Inspired by HyperMixer, we propose to adapt hypernetworks for the use in graphs.
Specifically, we propose \textit{HyperAggregation}, an aggregation function for GNNs that utilizes hypernetworks to aggregate arbitrarily sized neighborhoods, \ie neighborhoods with any number of vertices.
The hypernetworks are trained to predict dynamically shaped weights that fit the size of the current vertex's neighborhood.
Neighbor aggregation is then performed through matrix multiplication of the neighborhood's embeddings with the predicted weights.

We validate HyperAggregation with two models, one based on classical GNNs, called \textit{GraphHyperConv} (short: GHC), and the other one based on the MLP-Mixer, called the \textit{GraphHyperMixer} (short: GHM).
Experiments show that our models based on HyperAggregation perform well for vertex classification on homophilic and heterophilic graphs as well as on graph-level classification and regression tasks.
In most cases, GraphHyperConv outperforms GraphHyperMixer and reaches a new state of the art on the Roman-Empire dataset.
Overall our contributions are: 

\begin{itemize}
\item We propose HyperAggregation for GNNs that utilizes hypernetworks to generate neighborhood-depended weights and use them to aggregate neighborhood information.

\item We use HyperAggregation in two models: in a GCN-like model, called GraphHyperConv (GHC), where each layer aggregates over a $1$-hop neighborhood, and as an MLP-Mixer-like model called GraphHyperMixer (GHM).

\item We fairly evaluate both variants on a diverse set of benchmark tasks including vertex classification on homophilic and heterophilic datasets in both the transductive and inductive setup, graph classification, and graph regression.

\item An ablation study shows the effect of different hyperparameters, \eg the target network's hidden dimension.
\end{itemize}

Below, we summarize the related work.
Section~\ref{sec:methods} introduces our HyperAggregation approach.
The experimental apparatus is described in Section~\ref{sec:experimentalapparatus}.
We report and discuss the results of the experiments in Section~\ref{sec:res_disc}. 
Then we perform some ablation studies in Section~\ref{sec:ablation}, before we conclude.

\section{Related Work}
\label{sec:relatedwork}

\subsection{Hypernetworks}

Hypernetworks were first introduced by~\citet{DBLP:conf/iclr/HaDL17} to generate weights for Long Short-Term Memory and Convolutional Neural Network (CNN) models.
They use a layer embedding as input to a small hypernetwork which predicts weights for a bigger target model.
A recent survey~\citep{surv:hyp} lists possible use cases of hypernetworks as soft weight sharing, dynamic architectures, data-adaptive Neural Networks (NNs), uncertainty quantification, and parameter efficiency.

An application of hypernetworks in NLP is the HyperMixer~\citep{DBLP:conf/acl/MaiPFCMF023}.
It is designed as an efficient alternative to transformers and is based on the MLP-Mixer.
The MLP-Mixer~\citep{DBLP:conf/nips/TolstikhinHKBZU21} is a vision model that, like the Vision Transformer (ViT)~\citep{DBLP:conf/iclr/DosovitskiyB0WZ21}, uses image patches as input.
Each layer starts with channel mixing, \ie information is exchanged between all tokens by a multilayer perceptron (MLP) processing all tokens together channel by channel.
Then a token-mixing MLP follows which processes each token individually.
The HyperMixer predicts the weights of the channel mixing MLP with a hypernetwork.
This way the model is not limited to a fixed input size but can handle arbitrarily long input sentences, \ie it does not require the usual padding of short or cropping of long sentences. 

An unrelated though similarly named application of hypernetworks is the Graph HyperNetwork~\citep{DBLP:conf/iclr/ZhangRU19}.
It uses hypernetworks to speed up neural architecture search by utilizing the compute graph of a CNN to predict its weights.
In our work, hypernetworks are neural networks that predict weights for other networks and should not be confused with hypergraphs, \ie graphs where edges connect more than two vertices.
Thus, models like Hyper-SAGNN~\cite{DBLP:conf/iclr/ZhangZ020}, which can handle edges with a variable number of vertices, are not related to our work.

\subsection{Classical GNNs}

GCN~\citep{DBLP:conf/iclr/KipfW17} generalizes convolution to graphs.
To speed up training and inference, S-GCN~\citep{DBLP:conf/icml/WuSZFYW19} uses only a single GCN layer but with a higher power adjacency matrix.
Instead of stacking GCN layers, SIGN~\citep{SIGN/corr/abs-2004-11198} proposes to precompute features based on different powers of the adjacency matrix and the original features and to train an MLP on these new features.
SAGN~\citep{SAGN/corr/abs-2104-09376} improves on SIGN by using attention, self-labeling, and label propagation.
GCNII~\citep{DBLP:conf/icml/ChenWHDL20} modifies GCN to train deeper models among others by adding skip connections from the first to every layer.
\citet{DBLP:conf/icml/XuLTSKJ18} propose to add jumping knowledge connections, which connect every layer with the last one.
GAT~\citep{DBLP:conf/iclr/VelickovicCCRLB18} uses attention to give individual weights to neighbors.
GraphSAGE~\citep{DBLP:conf/nips/HamiltonYL17} and GraphSAINT~\citep{DBLP:conf/iclr/ZengZSKP20} focus on inductive performance, \ie performance on tasks where the test data is not used as structure information during training, and include different sampling strategies in their approaches.
Es-GNN~\citep{esgnn/corr/abs-2205-13700} focuses on heterophilic graphs, \ie graphs where neighbors usually do not share the same class, by partitioning the graph into a task-relevant and irrelevant part and aggregating information separately on each subgraph.
\citet{DBLP:conf/iclr/ZhaoJAS22} propose GNN-AK, which extends the aggregation step to use the whole induced subgraph rather than just a star like structure, as is common in message passing networks.

\subsection{Alternative GNNs}

There are a multitude of approaches that train MLPs on graphs.
Graph-MLP~\citep{graphMLP} uses a contrastive loss to pull neighboring vertices' embeddings closer together.
\citet{DBLP:conf/iclr/ZhangLSS22} motivate their Graph Less Neural Networks (GLNNs) by the need of fast inference time for deployment at scale.
GLNNs are MLPs that are trained through distillation of a classical GNN.
\citet{DBLP:conf/iclr/ZhengH0KWS22} also distill GNNs into MLPs with some tricks like adding special vertex structure embeddings to the teacher and trying to reconstruct them with the student MLP.
NOSMOG~\citep{DBLP:conf/iclr/0001ZG0C23} is another distillation procedure with tricks like concatenating DeepWalk~\citep{DBLP:conf/kdd/PerozziAS14} embeddings to the input of the MLP, a representation similarity loss, and adversarial feature noise.

With the current success of transformers in domains like NLP and vision, some works try to apply transformers and the idea of pretraining to graphs~\citep{graphbert,DBLP:conf/nips/YingCLZKHSL21,DBLP:conf/icml/ChenOB22}.
Graph-BERT~\citep{graphbert} is trained on a set of vertex' attributes for graph reconstruction tasks.
The approach was only tested on small citation graphs and trained from scratch for each dataset.
Graphormer~\citep{DBLP:conf/nips/YingCLZKHSL21} uses the default transformer architecture combined with a special structure encoding and a virtual vertex connected to all other vertices. 
They achieve competitive performance on large-scale benchmark datasets and utilize pretraining between compatible graphs.
\citet{DBLP:conf/icml/ChenOB22} utilize a classical GNN as subgraph encoder and input these encoded subgraphs into a transformer with a modified attention mechanism.
Finally, \citet{DBLP:conf/icml/HeH0PLB23} transfer the MLP-Mixer to graphs with their Graph ViT/MLP-Mixer. 
First, they create overlapping graph patches through graph partitioning.
Then they embed each graph partition or patch with a classical GNN and finally input those embeddings into a modified MLP-Mixer or ViT.
Thus, unlike our HyperAggregation, their model is not using hypernetworks.

\section{HyperAggregation}
\label{sec:methods}

\begin{figure*}
\centering
\includegraphics[width=0.90\textwidth]{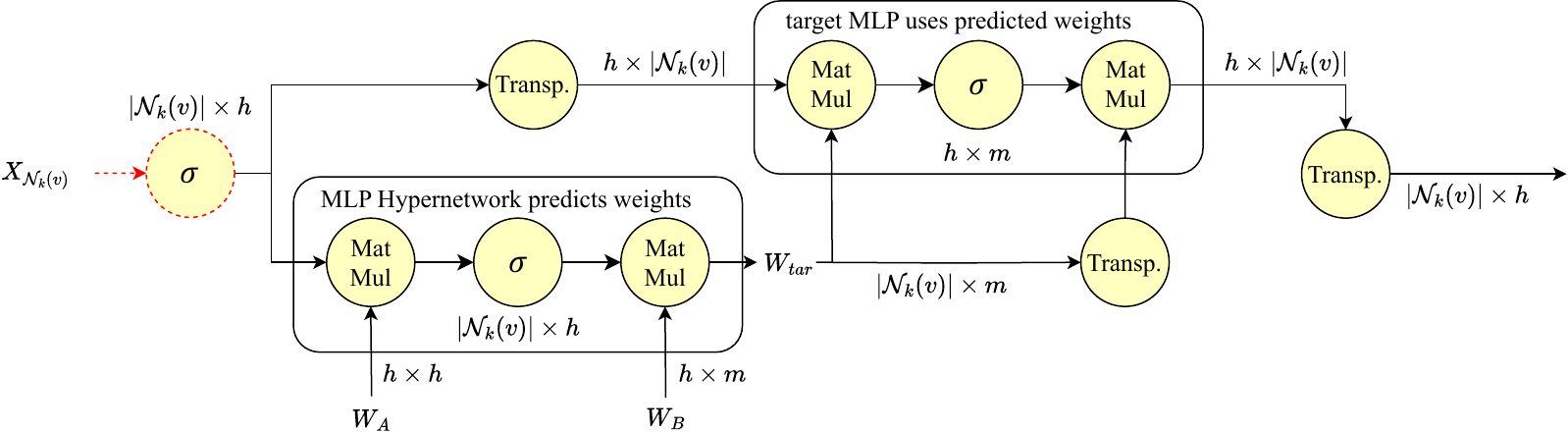}
\caption{Proposed HyperAggregation. The left MLP is the hypernetwork that predicts the target MLP's weights.
Its size depends on the vertex's neighborhood size. 
The target MLP uses those weights to perform the aggregation, \ie channel mixing across the neighborhood. Arrows are labeled with activation shapes. $W_A$ and $W_B$ are trainable parameters of the Hypernetwork. Dropout and normalization layers are not shown and the use of the leftmost activation $\sigma$ is optional.}
\label{fig:agg}
\end{figure*}

Following the recent applications of hypernetworks to replace the attention mechanism in transformers~\citep{DBLP:conf/acl/MaiPFCMF023}, we propose to use a hypernetwork to predict the target network's weights that then aggregates the neighborhood of a vertex.

First, we describe the basic idea of our aggregation function, then we demonstrate two different realizations of this aggregation function as GNNs.
We consider graphs $G=(V, E)$ with vertices $V$, which have features $X$, and edges $E$, which can also be represented as adjacency matrix $A$.
The vertex $v$'s $k$-hop neighborhood $\mathcal{N}_k(v)$ consists of all vertices that are connected to vertex $v$ with paths of at most length $k$.

\subsection{Aggregating using Hypernetworks}
A hypernetwork is a neural network with weights that are trained using back-propagation like any other standard neural network~\cite{DBLP:conf/iclr/HaDL17}.
The target network on the other hand has no trained weights.
Its weights are dynamically predicted by the hypernetwork depending on the current input to the model.
This allows us to dynamically adapt the target network's size to the current vertex's neighborhood that is aggregated.
We use two-layer MLPs as the hypernetwork and target network.

Figure~\ref{fig:agg} gives an overview of the aggregation function.
Normalization and dropout layers are hidden for clarity and the use of dashed red elements is optional and controlled by hyperparameters. 
Input to the aggregation is $X_{\mathcal{N}_k(v)} \in \mathbb{R}^{|\mathcal{N}_k(v)|\times h}$, the feature matrix of vertex $v$'s neighborhood with embedding dimension $h$.
After an optional activation function $\sigma$, this is then input into the MLP hypernetwork that has the trainable weights $W_A \in  \mathbb{R}^{h\times h}$ and $W_B \in  \mathbb{R}^{h\times m}$ with the mixing dimension $m$.
We always use the GeLU function as activation function $\sigma$. 
The hypernetwork predicts the target network's weight matrix $W_{tar} \in \mathbb{R}^{|\mathcal{N}_k(v)|\times m}$.
This shows how the target network's weights are dynamically generated to match the current neighborhood's size.
The original input $X_{\mathcal{N}_k(v)}$ is transposed to have shape ${h \times |\mathcal{N}_k(v)|}$ before it is input into the target MLP.
The target MLP then performs the neighborhood mixing using the predicted weights $W_{tar}$ in the first layer and the transposed predicted weights in the second layer.
In pre-experiments, we have confirmed that the setup of \citet{DBLP:conf/acl/MaiPFCMF023} to use one hypernetwork to predict $W_{tar}$ and use $W^T_{tar}$ as weights for the second layer improves the training stability compared to using two distinct hypernetworks.
Finally, the output of the target MLP is transposed back to the original shape  $|\mathcal{N}_k(v)|\times h$.

Formally, we define our HyperAggregation $HA$ as:
\begin{align*}
W_{tar} &= (\sigma(X_{\mathcal{N}_k(v)}W_A))W_B \\
HA(X_{\mathcal{N}_k(v)}) &= ((\sigma(X_{\mathcal{N}_k(v)}^TW_{tar}))W^T_{tar})^T
\end{align*}

Optionally, the activation function $\sigma$ can be applied to $X_{\mathcal{N}_k(v)}$ before putting it into the HyperAggregation.

\subsection{Models using HyperAggregation}

We test our proposed aggregation with two different architectures, called the GraphHyperMixer (GHM) and the GraphHyperConv (GHC).
The GraphHyperMixer is more similar to the Graph ViT/MLP-Mixer~\citep{DBLP:conf/icml/HeH0PLB23}.
The GraphHyperConv is similar to a classic GNN like GCN.

\begin{figure*}
    \centering
    \includegraphics[width=0.90\textwidth]{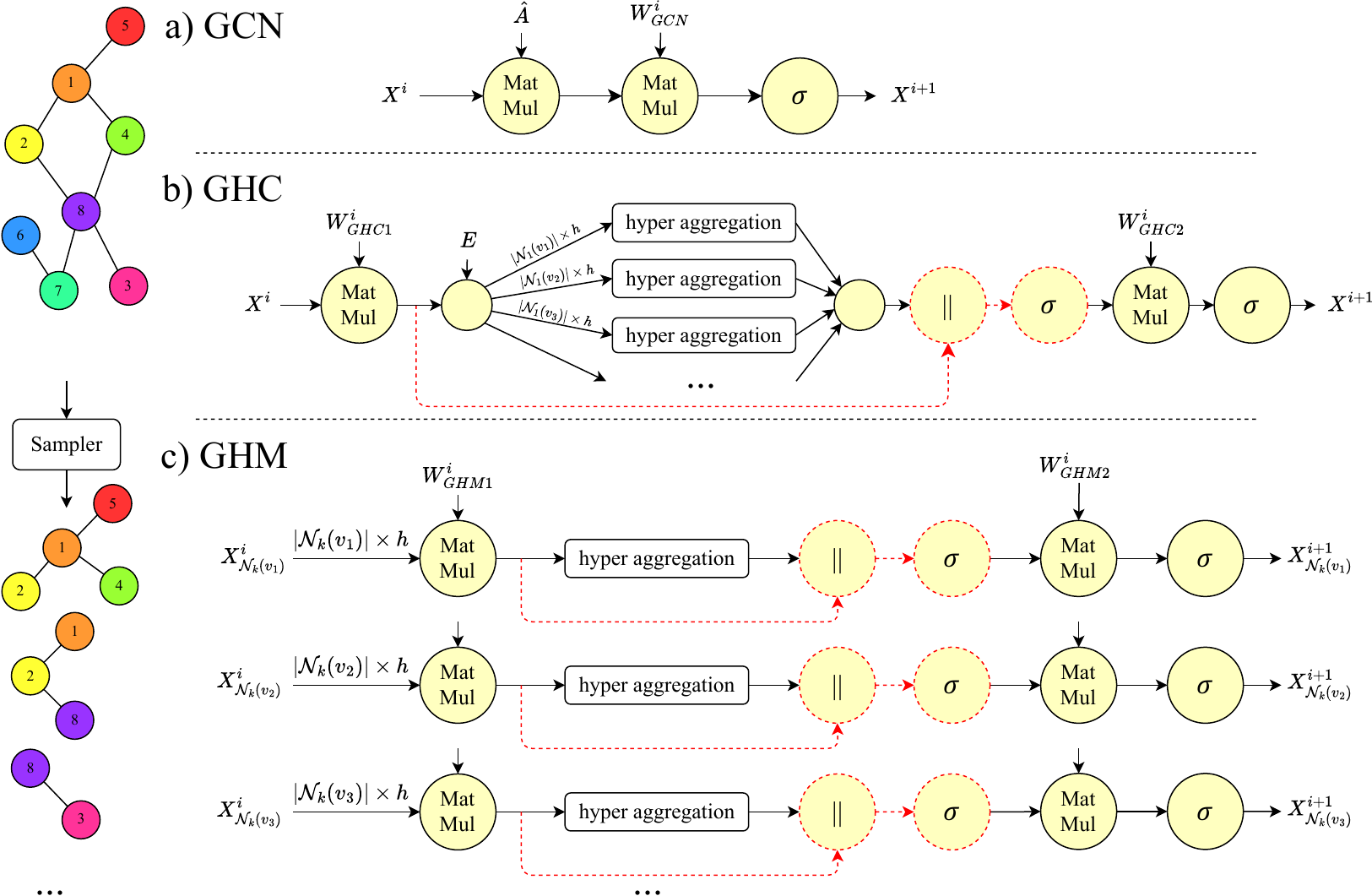}
    \caption{Depiction of a single a) GCN, b) GraphHyperConv, and c) GraphHyperMixer layer.}
    \label{fig:models}
\end{figure*}

Both architectures consist of multiple blocks, where each block consists of a Feed Forward (FF) layer, followed by the HyperAggregation (HA), followed by another FF layer.
The motivation of the two FF layers is that we can control the hidden size for the HyperAggregation even in a model with just one block.
In both models, the output of the HyperAggregation can be optionally concatenated with its input, which is commonly called a root connection, and passed through an activation before being fed into the final FF layer.
Figure~\ref{fig:models} shows a comparison of the two blocks and compares it with GCN.
Again, normalization and dropout layers are hidden for clarity and the use of dashed red elements is optional and controlled by hyperparameters. 

In the following, we highlight the differences between our two architectures using the HyperAggregation.

The \textit{GraphHyperMixer (GHM)} first samples the $k$-hop neighborhood $\mathcal{N}_k(v)$ of vertex $v$ with features $X_{\mathcal{N}_k(v)}^0$.
It assumes that the neighborhood $\mathcal{N}_k(v)$ is fully connected, \ie the aggregation treats all vertices in $\mathcal{N}_k(v)$ the same.
After multiple blocks, an output embedding for vertex $v$ is predicted.
In this setup, the blocks do not change the neighborhood size, so the receptive field and depth of the model are not dependent on each other.
Also, the aggregation's output is the same size as its input.
\begin{align*}
    X_\mathcal{N}^0 &= Sample(X, A) \\
    X_\mathcal{N}^{i+1} &= FF(HA(FF(X_\mathcal{N}^i)))
\end{align*}

The \textit{GraphHyperConv (GHC)} uses the whole graph as input.
In the multiple blocks, each vertex aggregates the information from its direct neighbors.
To obtain a single embedding after aggregating, depending on a hyperparameter, either the embedding of the root vertex or the mean over the neighborhood embeddings is used.
Similar to GCN and most other GNNs, the receptive field size depends on the number of blocks (or aggregations) used.
\begin{equation*}
X^{i+1} = FF(HA(FF(X^i), A))    
\end{equation*}

After introducing HyperAggregation and the two models we use it in, we will take a look at how we evaluate and compare its performance to other models.

\section{Experimental Apparatus}
\label{sec:experimentalapparatus}

\subsection{Datasets}
\label{sec:datasets}

\begin{table}
\centering    
\caption{Number of graphs, features, vertices, edges, classes, and number of training vertices / graphs for each dataset.}
\label{tab:dataset}
\resizebox{0.47\textwidth}{!}{
\begin{tabular}{l|rrrrrrr} \toprule
Dataset     & $|G|$ &$|X|$    &     $|V|$     &      $|E|$     & $|C|$ & \# train $V/G$  \\ \midrule
CiteSeer    &   $1$ &$3\,703$ &      $3\,327$ &       $9\,104$ &   $6$ &      $120$  \\
Cora        &   $1$ &$1\,433$ &      $2\,708$ &      $10\,556$ &   $7$ &      $140$  \\
PubMed      &   $1$ &   $500$ &     $19\,717$ &      $88\,648$ &   $3$ &       $60$  \\
OGB arXiv   &   $1$ &   $128$ &    $169\,343$ &  $1\,116\,243$ &  $40$ &  $90\,941$  \\ 
Computers   &   $1$ &   $767$ &     $13\,752$ &     $491\,722$ &  $10$ &      $300$  \\
Photo       &   $1$ &   $745$ &      $7\,650$ &     $238\,162$ &   $8$ &      $240$  \\ \midrule
Actor       &   $1$ &   $932$ &      $7\,600$ &      $30\,000$ &   $5$ &    $3\,648$  \\
Chameleon   &   $1$ &$2\,325$ &      $2\,277$ &      $36\,101$ &   $5$ &    $1\,092$  \\
Squirrel    &   $1$ &$2\,089$ &      $5\,201$ &     $217\,073$ &   $5$ &    $2\,496$  \\ 
Minesw.     &   $1$ &     $7$ &     $10\,000$ &      $39\,402$ &   $2$ &    $5\,000$  \\ 
Roman-E.    &   $1$ &   $300$ &     $22\,662$ &      $32\,927$ &  $18$ &   $11\,331$  \\ \midrule
MNIST       & $70\,000$ & $3$ &        $70.6$ &        $564.5$ &  $10$ &   $55\,000$\\
CIFAR10     & $60\,000$ & $5$ &       $117.6$ &        $941.2$ &  $10$ &   $45\,000$\\
ZINC        & $12\,000$ & $28$ &        $23.2$ &         $49.8$ &    N/A &   $10\,000$\\
\bottomrule
\end{tabular} }
\end{table}

We use various benchmark datasets as shown in Table~\ref{tab:dataset}.
The Cora~\citep{CoraCiteseer}, CiteSeer~\citep{CoraCiteseer}, PubMed~\citep{pubmed}, and OGB arXiv~\citep{ogb} 
datasets are citation graphs.
Computers~\citep{shchur} and Photo~\citep{shchur} are co-purchase graphs.
Following~\citet{shchur}, we use $10$ random splits with $20$ vertices per class for training, $30$ for validation, and all other vertices for testing for the datasets Cora, CiteSeer, PubMed, Computers, and Photo.
OGB arXiv uses papers before 2017 for training, papers from 2017 for validation, and papers after 2017 for testing.
The above graphs are homophilic, \ie neighboring vertices often share the same class.

The following graphs are more heterophilic, \ie neighboring vertices usually do not share the same class. 
Chameleon~\citep{ChamSqui} and Squirrel~\citep{ChamSqui} are Wikipedia page-page graphs and
Actor~\citep{geomgcn} is a Wikipedia co-occurrence graph for which we use the common $48\%/ 32\%/ 20\%$ splits~\citep{geomgcn}.
\citet{heter} critique some commonly used heterophilic datasets for low diversity, class imbalance, and duplicate vertices.
They introduce multiple new datasets, among them the text-based graph Roman-empire and the synthetic Minesweeper graph, which we use with their provided $10$ random splits.

Additionally, we use datasets with graph-level tasks.
The MNIST and CIFAR10 datasets~\cite{DBLP:journals/jmlr/DwivediJL0BB23} represent super-pixel, \ie homogeneous regions of the original image, as vertices.
The ZINC dataset~\cite{DBLP:journals/jcisd/SterlingI15,DBLP:journals/corr/Gomez-Bombarelli16} is a molecule dataset with the regression task to predict the constrained solubility.
We use a subset of the full dataset with $10\,000$ graphs for training and $1\,000$ graphs each for validation and testing \cite{DBLP:journals/jmlr/DwivediJL0BB23}.

\begin{table*}
\centering
\caption{Mean accuracy and standard deviation for the homophilic datasets.}
\label{tab:results_ho}
\begin{tabular}{l|lll|rr|r} \toprule
& Cora & CiteSeer & PubMed & Comp. & Photo & arXiv  \\ \midrule
MLP & $56.29_{2.08}$ & $55.54_{3.47}$ & $68.27_{2.29}$ & $67.43_{1.79}$ & $78.75_{1.73}$ & $57.62_{0.07}$ \\  \midrule \midrule
\textbf{transductive} \\ 
GCN & $78.43_{0.85}$ & $66.75_{1.86}$ & $75.62_{2.24}$ & $\mathbf{83.52_{1.73}}$ & $90.94_{1.06}$ & $\mathbf{72.52_{0.37}}$ \\ 
GCN~\cite{shchur} & $81.5_{1.3}$ & $\mathbf{71.9_{1.9}}$ & $77.8_{2.9}$ & $82.6_{2.4}$ & $91.2_{1.2}$ & N/A \\
GAT~\cite{shchur} & $\mathbf{81.8_{1.3}}$ & $71.4_{1.9}$ & $\mathbf{78.7_{2.3}}$ & $78.0_{19.0}$ &  $85.7_{20.3}$ & N/A \\
SAGE~\cite{shchur} & $79.2_{7.7}$ & $71.6_{1.9}$ & $77.4_{2.2}$ & $82.4_{1.8}$ & $91.4_{1.3}$ & N/A \\
\midrule
GHM & $77.33_{2.45}$ & $67.12_{1.41}$ & $75.80_{3.69}$ & $80.73_{1.76}$ & $88.56_{1.44}$ & $70.33_{0.24}$ \\ 
GHC & $78.85_{2.14}$ & $66.82_{1.66}$ & $76.31_{2.71}$ & $82.12_{1.91}$ & $\mathbf{91.63_{0.79}}$ & $72.41_{0.15}$ \\ \midrule \midrule
\textbf{inductive} \\
GCN & $75.19_{3.20}$ & $\mathbf{67.02_{1.34}}$ & $\mathbf{75.84_{2.56}}$ & $\mathbf{80.43_{1.97}}$ & $88.93_{0.87}$ & $\mathbf{72.67_{0.19}}$ \\
GHM & $75.44_{2.49}$ & $65.44_{2.78}$ & $75.15_{2.95}$ & $75.54_{3.17}$ & $\mathbf{89.83_{1.89}}$ & $69.59_{0.23}$ \\ 
GHC & $\mathbf{76.69_{1.61}}$ & $64.10_{2.55}$ & $73.28_{2.70}$ & $78.79_{2.26}$ & $89.70_{1.30}$ & $72.46_{0.23}$ \\ \bottomrule
\end{tabular}
\end{table*}

\begin{table*}
\centering
\caption{Mean accuracy and standard deviation for the heterophilic datasets. Best Specialized denotes the highest number of any of the models specialized for heterophilic data from~\citet{heter}.}
\label{tab:results_he}
\begin{tabular}{l|rrr|rr} \toprule
& Cham. & Squirrel & Actor & Minesweeper & Roman-Empire  \\ \midrule
MLP & $45.57_{1.77}$ &$33.35_{1.60}$ & $34.90_{0.81}$ & $80.00_{0.00}$ & $66.65_{0.47}$ \\  \midrule \midrule
\textbf{transductive} \\ 
GCN  & $69.63_{2.41}$ & $59.56_{1.92}$ & $33.70_{1.26}$ & $\mathbf{88.72_{0.52}}$  & $82.72_{0.82}$ \\ 
GCN~\cite{heter} & $50.18_{3.29}$ & $39.06_{1.52}$ & N/A & N/A & $73.69_{0.74}$ \\
SAGE~\cite{heter} & $50.18_{1.78}$ & $35.83_{1.32}$ & N/A & N/A & $85.74_{0.67}$  \\
GAT~\cite{heter} & $45.02_{1.75}$ & $32.21_{1.63}$ & N/A & N/A & $80.87_{0.30}$ \\
Best Specialized~\cite{heter} & $\mathbf{77.85_{0.46}}$ & $\mathbf{68.93_{1.69}}$ & N/A & N/A & $79.92_{0.56}$ \\
\midrule
GHM & $69.65_{2.27}$ & $63.49_{1.06}$ & $\mathbf{37.56_{1.26}}$ & $84.02_{0.77}$ & $83.97_{0.50}$ \\ 
GHC & $74.78_{1.82}$ & $62.90_{1.47}$ & $36.40_{1.46}$ & $87.49_{0.61}$ & $\mathbf{92.27_{0.57}}$\\ \midrule \midrule
\textbf{inductive} \\
GCN & $\mathbf{61.51_{3.80}}$ & $47.58_{2.46}$ & $34.26_{1.33}$ & $\mathbf{86.40_{0.56}}$ & $78.34_{0.57}$  \\
GHM & $59.71_{2.52}$ & $47.42_{2.44}$ & $\mathbf{36.64_{0.54}}$  & $83.88_{0.39}$ & $80.12_{0.46}$  \\ 
GHC & $60.68_{1.63}$ & $\mathbf{55.21_{2.59}}$ & $35.05_{1.51}$ & $85.23_{0.79}$ & $\mathbf{86.23_{0.62}}$  \\ \bottomrule
\end{tabular}
\end{table*}

\subsection{Procedure}
\label{sec:procedure}
For Cora, CiteSeer, PubMed, Computers, and Photo, we generate $10$ random splits and store them such that each model is trained on the same random splits.
For the other datasets, we use the provided splits.
On the larger datasets MNIST, CIFAR10, ZINC, and OGB arXiv we run the experiments with three different seeds.
On all other datasets, we run the experiment with $10$ different seeds.

We use both the transductive and inductive settings for the vertex-level tasks.
In the transductive setting, the whole graph is available for training, but the labels are restricted to only those of the training vertices.
In the inductive setting, only the training vertices with labels and the edges connecting two training vertices are available.
For the datasets that use $20$ training vertices per class, the inductive setting leads to very sparse training graphs.
Depending on the random sample, it can result in a completely disconnected graph with no edges.
To alleviate this issue, we follow the setup of \citet{NOSMOG}, which splits the test data into $80\%$ unlabeled vertices that can be used during training and $20\%$ test vertices.

For the graph-level tasks, we add a mean-pooling layer that pools the vertex embeddings before the final classification layer. 
For a fairer comparison, we do not use the edge attributes of the ZINC dataset, as our baseline models can not utilize them.
All graph-level tasks are inductive, as each graph is either in the train, validation, or test set and there are no edges between distinct graphs.

We compare GraphHyperMixer and GraphHyperConv with the baselines GCN and MLP as well as results from the literature.
Note that the MLP only uses the training vertices and thus is always inductive.
For the homophilic datasets, we use the GCN, GAT, and GraphSAGE results from \cite{shchur}.
For the heterophilic datasets, we compare against the same models and additionally use the best of the heterophilic optimized models from \cite{heter}.
For the graph-level datasets we compare against the same 4-layer models from \cite{DBLP:journals/jmlr/DwivediJL0BB23}.
For ZINC, we use their results that did not utilize edge features to keep the comparison fair.
We report test accuracy for all experiments except for ZINC, where we report the mean absolute error (MAE).

\subsection{Hyperparameter}

We optimize the GNN hyperparameters through a two-step grid search. 
First, we optimize all architectural hyperparameters while using only two values for dropout and weight decay each with a reduced number of repeats. 
These include whether to include reverse and self-edges in the datasets (if they do not include them by default), model depth, hidden size, learning rate, and whether to use residual connections.
Some models also use sampling and therefore have the additional hyperparameters of a batch size and sampled subgraph size.
For the models using HyperAggregation, we additionally tune the mixing dimension, whether to use a skip connection that brings the unmodified root embedding over the aggregation, whether to use layer norm and dropout before the target network, and whether to use layer norm and dropout after the target network.
For GraphHyperConv, we additionally determine if it is better to use just the root vertex embedding after aggregation or the mean-pool over the neighborhood.
In a second step with all above hyperparameters fixed we optimize the input dropout, model dropout, mixing dropout if applicable, and weight decay.
The implementation and all final hyperparameter values to reproduce our results are available under \url{https://github.com/Foisunt/HyperAggregation}.

\section{Results and Discussion}
\label{sec:res_disc}

Table~\ref{tab:results_ho} shows the results of our experiments on homophilic datasets. 
In the transductive setup, GHC is the best model on Photo, while in the inductive setup, GHM is the best model on Photo and GHC on Cora.
Table~\ref{tab:results_he} shows the results of our experiments on heterophilic datasets. 
In the transductive case on the Chameleon and Squirrel datasets, the best model from the comparison study of heterophilic models by~\citet{heter} outperform our GHM and GHC.
On Actor and Roman-Empire GHM and GHC perform best, respectively.
Averaged over all vertex level tasks, GHM has $ 70.79\%$ accuracy which is worse than GCN's $71.38\%$.
GHC beats the other two models with an average of $72.70\%$. 
Table~\ref{tab:results_gra} shows the results of our experiments on graph-level datasets.
GraphSAGE performs best on the graph classification datasets and GHC performs best on the graph regression task of the ZINC dataset.

\begin{table}
\centering
\caption{Results on the graph-level tasks. Accuracy for MNIST and CIFAR10 and MAE on ZINC.}
\label{tab:results_gra}
\begin{tabular}{l|rrr} \toprule
& MNIST $\uparrow$ & CIFAR10 $\uparrow$ & ZINC $\downarrow$   \\ \midrule
MLP & $92.53_{0.27}$ & $52.28_{0.12}$ & $0.731_{0.000}$ \\   
GCN & $90.60_{0.22}$ & $49.74_{0.29}$ & $0.426_{0.013}$ \\ 
GCN~\cite{DBLP:journals/jmlr/DwivediJL0BB23} & $90.12_{0.15}$ & $54.14_{0.39}$ & $0.416_{0.006}$ \\ 
GAT~\cite{DBLP:journals/jmlr/DwivediJL0BB23} & $95.54_{0.21}$ & $64.22_{0.46}$ & $0.475_{0.007}$  \\ 
SAGE~\cite{DBLP:journals/jmlr/DwivediJL0BB23} & $\mathbf{97.31_{0.10}}$ & $\mathbf{65.77_{0.31}}$ & $0.468_{0.003}$ \\ 
\midrule
GHM & $96.74_{0.18}$ & $53.30_{0.31}$ & $0.669_{0.003}$\\ 
GHC & $97.24_{0.12}$ & $59.83_{0.61}$ & $\mathbf{0.337_{0.020}}$ \\ \bottomrule
\end{tabular}
\end{table}

\paragraph*{GHC usually outperforms GHM}
When studying the result tables, we see that GHC outperforms GHM in many cases.
The main exceptions are the CiteSeer and Actor datasets, where GHM beats GHC in both settings.
The reason is probably that GHM ignores the edge information after the graph is split into subgraphs and assumes each of these subgraphs is fully connected.
This leads to a trade-off between giving GHM enough context by choosing large enough subgraphs versus not knowing the exact structure of the subgraph.
When using a $1$-hop neighborhood there might be not enough features to classify each vertex correctly.
When using a many-hop neighborhood, we have more vertex features, but we lose the structural information in the GHM as it does not rely on the adjacency matrix.
In contrast, GHC does not suffer from this issue as it, like most other GNNs, utilizes edge information in every aggregation layer to know the exact neighborhoods. 
It is interesting though that GHM only outperforms GHC in the inductive setting, so we take a closer look at that.

\paragraph*{Transductive vs Inductive}

In general, one expects the performance to be higher in the transductive setting as more information about the test data is available during training than in the inductive setting.
We can observe this to a varying degree between the different models.
GCN's performance decreases by an average of $3.08$ points from $72.92$ in the transductive setting to $69.83$ in the inductive setting.
The performance of GHM drops by an average of $3.62$ points from $72.60$ to $68.98$.
GHC's performance drops from $74.73$ to $70.67$ by $4.05$ points. 
This shows that GHC performs especially well in the transductive setting with more information available.
Although GHC loses performance when switching to the inductive setting, it still stays the best model in this comparison.
When we look at the performance difference on a per-dataset level, we can see that most models behave similarly on most datasets.
For example, on OGB arXiv, there is only a minor performance difference between the inductive and transductive setup, while on the heterophilic Chameleon and Squirrel datasets we observe differences of over $10$ points.

\paragraph*{Homophilic vs Heterophilic Graphs}
Our results in Tables~\ref{tab:results_ho} and~\ref{tab:results_he} demonstrate that HyperAggregation works well for both homophilic and heterophilic datasets.
On Chameleon and Squirrel, the best model identified in the comparison study of \citet{heter} is also the best in our comparison.
However, it has to be noted that the analysis of \citet{heter} already concludes that the two datasets Chameleon and Squirrel are problematic and require a cleanup.
Thus, the authors suggest new heterophilic datasets.
For the Roman-Empire datasets, the best models fall behind baselines such as GCN and GAT.

GHC shows good performance on all heterophilic datasets but performs especially well on Roman-Empire.
The comparison paper~\cite{heter} did not use Actor and reported the AUROC score for Minesweeper instead of accuracy, therefore these numbers are missing from Table~\ref{tab:results_he}.
On Roman-Empire, GHC's $92.27\%$ accuracy is the new state of the art as far as we are aware.
It outperforms the best results from \citet{heter}, which are a modified graph transformer achieving $87.32\%$ and a modified GraphSAGE achieving $88.75\%$, as well as CO-GNN($\Sigma$, $\Sigma$)~\cite{finkelshtein2023cooperative} achieving $91.57\%$, DIR-GNN~\cite{rossi2023edge} with $91.23\%$, and CDE-GRAND~\cite{zhao2023graph} achieving $91.64\%$.

\paragraph*{Graph-level Tasks}
The results in Table~\ref{tab:results_gra} show that our HyperAggregation performs well on graph-level tasks.
On the graph classification datasets MNIST and CIFAR10, GHM and GHC perform similar to comparison models.
Compared to all other datasets, MLP performs on par with GCN on these two datasets.
A reason for this might be that MLP has access to the features of the whole graph through the final pooling layer.
The results on ZINC are quite interesting as GHM performs quite poorly, but GHC is the best model in our comparison.
However, GHC's lead does not hold when comparing it to the results of models that utilize the edge features or are much bigger like the $0.278$ MAE of the 16-layer GCN from \citet{DBLP:journals/jmlr/DwivediJL0BB23}.

\paragraph*{Assumptions and Limitations}
\label{sec:threattovalidity}
GraphHyperConv assumes that there is at least one vertex in each neighborhood.
This is a limitation of our implementation within the pytorch geometric framework.
If the input graph has unconnected vertices or some vertices become unconnected in the inductive setting, this can be remedied by adding self-loops to the graph.
For GraphHyperMixer, this is not an issue as the root vertex is always part of the subgraph sampled at the beginning and the aggregation assumes a fully connected subgraph.

\begin{figure*}
 \centering
 \begin{subfigure}[b]{0.49\textwidth}
     \centering
     \includegraphics[width=\textwidth]{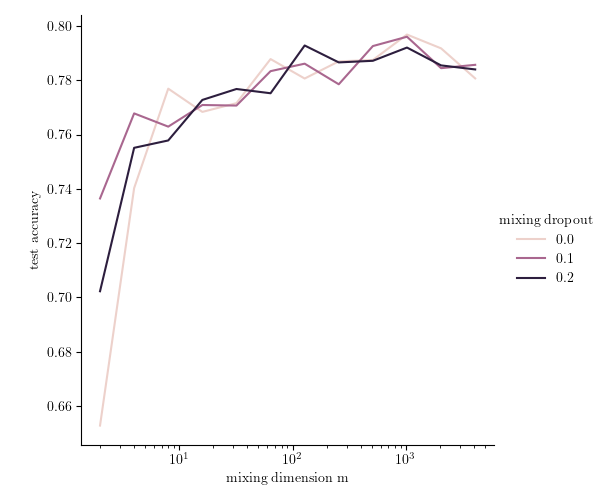}
     \caption{Ablation on the mixing dimension of GHC on Cora.}
     \label{fig:abl_ghc_mixdim_cora}
 \end{subfigure}
 \hfill
 \begin{subfigure}[b]{0.49\textwidth}
     \centering
     \includegraphics[width=\textwidth]{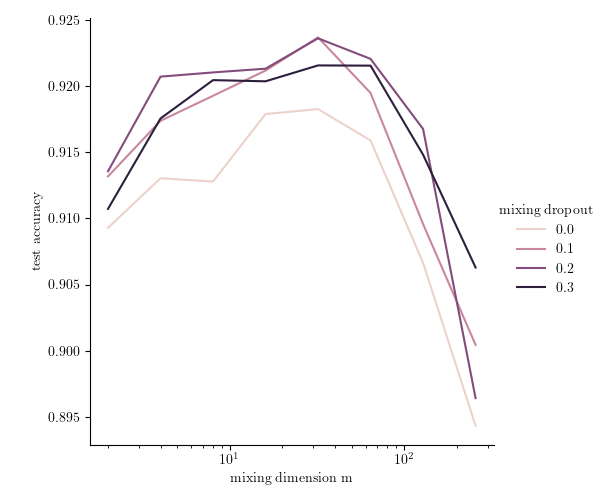}
     \caption{Ablation on the mixing dimension of GHC on Roman-Empire.}
     \label{fig:abl_ghc_mixdim_re}
 \end{subfigure}
 \hfill
 \begin{subfigure}[b]{0.49\textwidth}
     \centering
     \includegraphics[width=\textwidth]{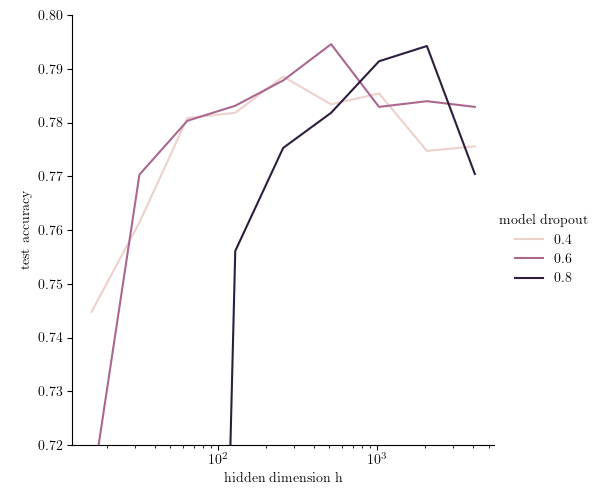}
     \caption{Ablation on the hidden dimension of GHC on Cora.}
     \label{fig:abl_ghc_hidden_cora}
 \end{subfigure} 
 \hfill
 \begin{subfigure}[b]{0.49\textwidth}
     \centering
     \includegraphics[width=\textwidth]{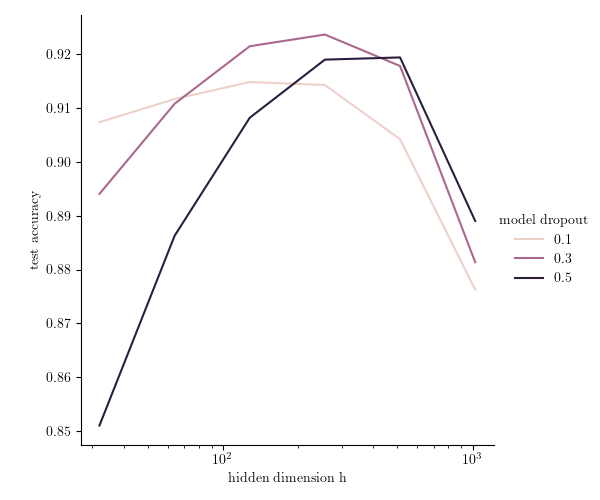}
     \caption{Ablation on the hidden dimension  of GHC on Roman-Empire.}
     \label{fig:abl_ghc_hidden_re}
 \end{subfigure}
    \caption{Ablation studies of the mixing and hidden dimension of GraphHyperConv.}
    \label{fig:three graphs}
\end{figure*}

\section{Ablation Studies}
\label{sec:ablation}

To validate the robustness of HyperAggregation and GHC w.r.t. the chosen hyperparameters, we perform various ablation studies.
Note that all ablation studies report test accuracy.
This means that our reported numbers of the main result tables can be lower than the highest values shown here, as our hyperparameter search on validation accuracy might not have found these optimal values.

\paragraph*{Ablation on Mixing and Hidden Dimension}

Figures~\ref{fig:abl_ghc_mixdim_cora} and \ref{fig:abl_ghc_mixdim_re} show the test accuracy when changing the mixing dimension $m$, \ie the hidden dimension of the target network of GraphHyperConv on Cora and Roman-Empire, respectively.
In the main experiments, we use $m=64$ without mixing dropout and $m=32$ with a mixing dropout of $0.1$ for Cora and Roman-Empire, respectively.
The mixing dropout is applied to the vertex embeddings that are input to the target network.
\citet{DBLP:conf/acl/MaiPFCMF023} used a mixing dimension of $512$. 
Our ablations show that, depending on the dataset, an optimal mixing dimension is notably smaller, \eg $32$ for Roman-Empire.
On Cora, Figure~\ref{fig:abl_ghc_mixdim_cora} shows that, depending on the dropout setting, once a mixing dimension of about $128$ is reached, raising the dimension further only minimally increases performance.
For Roman-Empire, Figure~\ref{fig:abl_ghc_mixdim_re} shows that $32$ is the best mixing dimension independent of the dropout setting.
Unlike on Cora, increasing the mixing dimension further does not provide any benefits here.

Figures~\ref{fig:abl_ghc_hidden_cora} and \ref{fig:abl_ghc_hidden_re} show the performance when changing the hidden dimension in our model, except for the target network.
Figure~\ref{fig:abl_ghc_hidden_cora} shows that increasing the hidden dimension improves performance on Cora.
On Cora, our hyperparameter search found a hidden dimension of $256$ and a model dropout of $0.6$ to perform best on the validation set.
On Roman-Empire, we used a hidden dimension of $256$ and a model dropout of $0.3$.
The figures show that both a very low and very high hidden dimension reduces performance on either dataset.
In general, these ablations show that GHC is quite stable to reasonable hidden dimensions, but also that a larger hidden dimension for Cora could slightly improve the results.

\paragraph*{Ablation on Architectural Hyperparameters}
Table~\ref{tab:abl} shows the effect of different hyperparameter choices for GHC.
Removing self-loops on Cora and adding them for Roman-Empire decreases performance.
Not making Roman-Empire undirected decreases performance.
Cora is already undirected and thus it is not applicable.
Adding input normalization reduces performance on Cora while dropping it for Roman-Empire only has a marginal effect.
Adding residual connections to GHC on Cora or dropping them on Roman-Empire both decreases the performance.
The root connection, \ie concatenating the original root vertex's embedding with the output of the HyperAggregation, gives a consistent boost to the performance on both datasets.
Using the mean of the neighborhood instead of just the root vertex's embedding after aggregating improves performance on Cora but decreases performance on Roman-Empire.
Finally, the better place to use dropout, an activation function, and layer norm is after HyperAggregation for Cora but before HyperAggregation for Roman-Empire.
Combined with the first ablations, we can state that our models are quite robust against many hyperparameter choices.

As default settings we recommend using the root connection, to not normalize the inputs of the model, and to add layer norm and dropout after the aggregation.
Whether to use residual connection depends on the dataset and depth of the model.
Finally, testing a few hidden and mixing dimensions as well as dropout settings will lead to good results.

\begin{table}
\centering
\caption{Ablation on other hyperparameters of GHC, all ablations are compared to the base performance and only change the mentioned hyperparameter.}
\begin{tabular}{l|rr|rr} \toprule
  & \multicolumn{2}{c|}{Cora}  & \multicolumn{2}{c}{Roman-Empire} \\
Hyperparameter   & changed to & accuracy & changed to & accuracy \\ \midrule
Base performance & N/A & $78.85$ & N/A & $92.27$ \\ \midrule
Make undirected  & N/A & N/A & no & $-5.89$ \\
Add self loops   & no  & $-2.84$ & yes & $-0.13$ \\ \midrule
Normalize input  & yes & $-0.66$ & no & $-0.01$  \\
Residual connections & yes & $-3.09$ & no & $-1.22$ \\ 
Root connection  & no & $-0.50$ & no & $-1.72$ \\ 
Mean aggregate   & no & $-1.35$ & yes & $-2.64$ \\ 
Trans HA input   & no &  $0.47$ & no &  $-1.15$ \\ 
Trans HA output  & no & $-4.83$ & yes & $-0.08$  \\ 
\bottomrule
\end{tabular}
\label{tab:abl}
\end{table}

\section{Conclusion and Future Work}
\label{sec:conclusion}
We have introduced HyperAggregation, an aggregation function over graph edges based on hypernetworks.
We demonstrated the use of our HyperAggregation in two ways.
GraphHyperConv uses HyperAggregation to replace a classical aggregation function, while GraphHyperMixer is an MLP-Mixer model for graphs.
Experiments show the effectiveness of the approaches for vertex classification on homophilic and heterophilic graphs as well as graph-level classification and regression.
GraphHyperConv achieves a new state of the art on the heterophilic Roman-Empire dataset.
We also demonstrated its robustness against certain hyperparameter choices in detailed ablation studies.

Our work shows the principal applicability of a hypernetwork-based aggregation function for graphs.
This opens future works such as combining HyperAggregation with other techniques and studying its behavior. 
With GraphHyperConv, we have shown that the HyperAggregation can be used as a generic aggregation function.
That means any tricks that have been developed for other GNNs can be applied together with HyperAggregation to improve the performance for some use cases.
For example, more informative features as in GIANT~\citep{DBLP:conf/iclr/ChienCHYZMD22} and SIGN~\citep{SIGN/corr/abs-2004-11198}, label reuse~\citep{DBLP:journals/corr/abs-2103-13355}, label/prediction propagation~\citep{SAGN/corr/abs-2104-09376,DBLP:conf/ijcnn/HoffmannGS23}, modifying the adjacency matrix for heterophilic tasks~\citep{esgnn/corr/abs-2205-13700}, and other tricks can be added.
Another avenue for future work would be the utilization of edge features.
For GHC this could be achieved by concatenating the edge features to $X_{\mathcal{N}_k(v)}$ as input to the HyperAggregation.

\section*{Acknowledgment}
This work was performed on the computational resource bwUniCluster funded by the Ministry of Science, Research and the Arts Baden-Württemberg and the Universities of the State of Baden-Württemberg, Germany, within the framework program bwHPC.

\printbibliography

\end{document}